\documentclass[letterpaper]{article} 
\usepackage{aaai25}  
\usepackage{times}  
\usepackage{helvet}  
\usepackage{courier}  
\usepackage[hyphens]{url}  
\usepackage{graphicx} 
\usepackage{natbib}  
\usepackage{caption} 
\usepackage{algorithm}
\usepackage{algorithmic}

\usepackage{newfloat}
\usepackage{listings}
\usepackage{booktabs}       
\usepackage{comment}
\usepackage{graphicx}
\usepackage{caption}
\usepackage{subcaption}

\DeclareCaptionStyle{ruled}{labelfont=normalfont,labelsep=colon,strut=off} 
\floatstyle{ruled}
\newfloat{listing}{tb}{lst}{}
\floatname{listing}{Listing}
\title{The Unreasonable Effectiveness of Open Science in AI: A Replication Study}

\author {
    Odd Erik Gundersen\textsuperscript{\rm 1,2},
    Odd Cappelen\textsuperscript{\rm 1,3},
    Martin Mølnå\textsuperscript{\rm 1,2},
    Nicklas Grimstad Nilsen\textsuperscript{\rm 1,2}
}
\affiliations {
    \textsuperscript{\rm 1}Norwegian University of Science and Technology, Trondheim, Norway\\
    \textsuperscript{\rm 2}Aneo AS, Trondheim, Norway\\
    \textsuperscript{\rm 3}Minus 1, Oslo, Norway\\
    odderik@ntnu.no,
    odd.cappelen@minus1.no,
    \{martin.molna,nicklas.nilsen\}@aneo.com, 
}

\begin{document}

\maketitle

\begin{abstract}
    A reproducibility crisis has been reported in science, but the extent to which it affects AI research is not yet fully understood.
    Therefore, we performed a systematic replication study including 30 highly cited AI studies relying on original materials when available.
    In the end, eight articles were rejected because they required access to data or hardware that was practically impossible to acquire as part of the project. 
    Six articles were successfully reproduced, while five were partially reproduced.
    In total, 50\% of the articles included was reproduced to some extent. 
    The availability of code and data correlate strongly with reproducibility, as 86\% of articles that shared code and data were fully or partly reproduced, while this was true for 33\% of articles that shared only data.
    The quality of the data documentation correlates with successful replication.
    Poorly documented or miss-specified data will probably result in unsuccessful replication.
    Surprisingly, the quality of the code documentation does not correlate with successful replication. 
    Whether the code is poorly documented, partially missing, or not versioned is not important for successful replication, as long as the code is shared.
    This study emphasizes the effectiveness of open science and the importance of properly documenting data work.
\end{abstract}



%


\section{Introduction}
There is evidence that science is undergoing a reproducibility crisis \cite{baker2016reproducibility}. 
\cite{ioannidis2005most} claims that most published research findings are false. 
Research in artificial intelligence (AI) is not spared \cite{hutson2018artificial,haibe2020transparency,gundersen2020reproducibility}, and machine learning-based science, which is science that uses machine learning algorithms, is severely affected \cite{kapoor2023leakage}.
Some even argue that AI might be one of the causes of irreproducibility \cite{ball2023ai}.
However, it is not clear how much science is reproducible. 
Notable examples of studies that have tried to estimate the reproducibility exist for psychology \cite{open2015estimating}, social science \cite{camerer2018evaluating}, economics \cite{camerer2016evaluating}, medicine \cite{prinz2011believe}, and social psychology \cite{klein2014investigating}.
\cite{raff2019step} estimated the reproducibility of machine learning to be 63.5\% by attempting to independently reproduce 255 articles by reimplementing the algorithms, while \cite{gundersen2018state} estimated the reproducibility of research published in AAAI and IJCAI to be 26\% indirectly by analyzing 400 research articles.

\begin{figure}[t!]
\centering
\includegraphics[width=1\columnwidth]{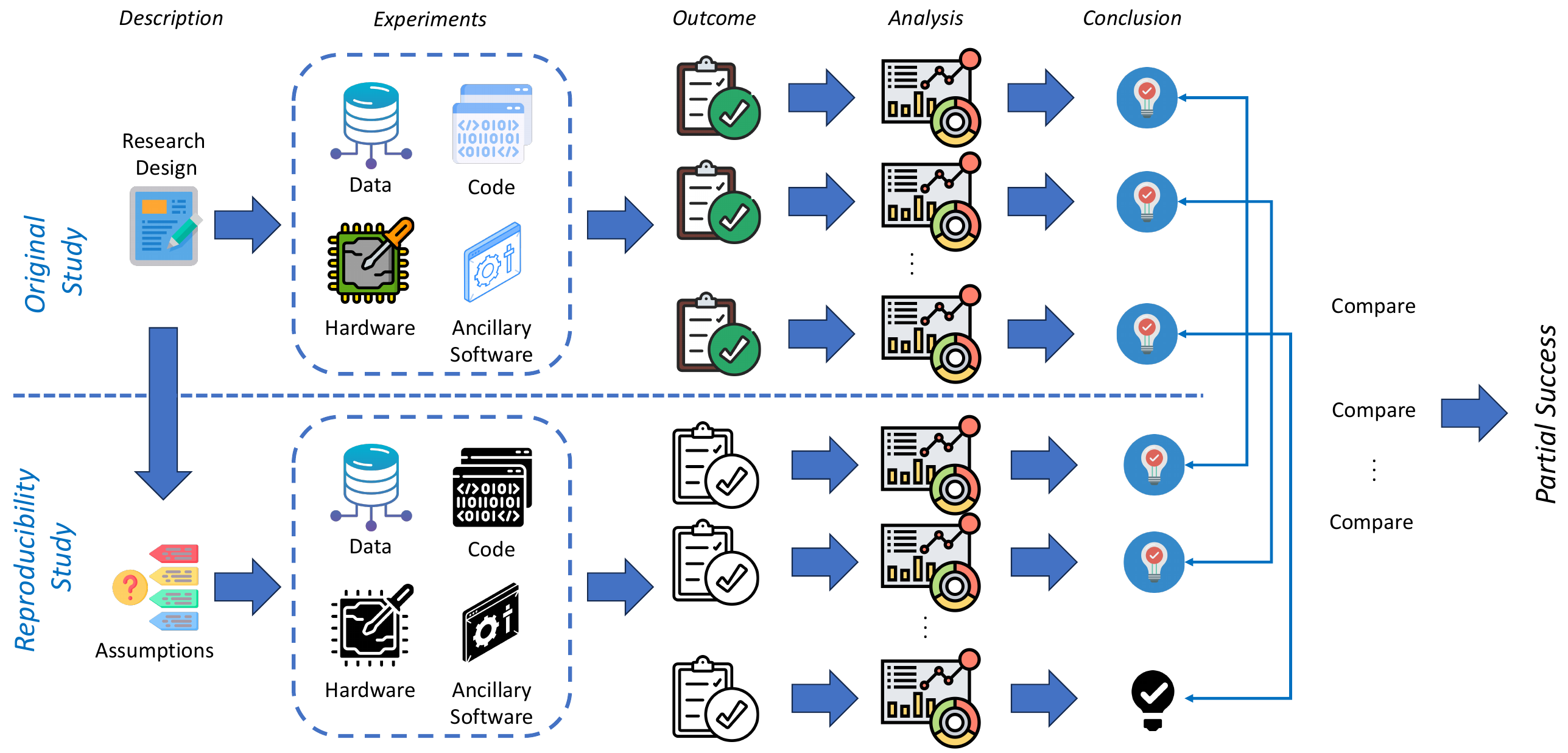} 
\caption{
An R3 type reproducibility study is conducted where only the research report and data are shared publicly, so the code implementing both the AI method and experiment as well as hardware an ancillary software differ from the original experiment, as illustrated by differing icons. 
Many assumptions must be made when conducting a reproducibility study as the research report cannot possibly cover all design decisions.
Difference in implementation as well as differences in hardware and ancillary software can introduce differences between the output produced by the original and the reproducibility experiments. 
Sometimes the differences lead to different conclusions  even after doing the exact same analysis. 
This is indicated for one of the reproducibility experiments by the black light bulb icon. 
When the original article describes several experiments, the overall result of the reproducibility study depends on whether all the conclusions of all these experiments agree or not. 
If only a subset of the conclusions is the same, the reproducibility study is considered \emph{Partial Success}.}
\label{fig:reproducibility_experiment}
\end{figure}

Many studies in machine learning have investigated reproducibility in various ways. 
\cite{pham_2020} and \cite{zhuang_2022} investigate how different sources of variability affect the results of neural networks, and \cite{gundersen2023reporting} investigate how variability affects conclusions.
Natural language processing is the focus of \cite{melis_2018}, who find that standard LSTM architectures, when properly regularized, outperform more recent models, and \cite{belz_2021a} found that replication under the same conditions yields results that are more often worse than results that are better than those reported.
\cite{lucic_2018} 
did not find any evidence that the algorithms they tested consistently outperformed \cite{goodfellow2014generative}, although this was claimed in the articles introducing these algorithms.
\cite{bouthillier_2019} focus on image recognition. 
They note the lack of clarity on whether exploratory or confirmatory research is reported and argue the need to increase the rigor of empirical research in deep learning. 
\cite{varoquaux_2022} argue that as long as there is no incentive for clarity in publications, researchers optimize for publication. 
Therefore,  publication norms should be improved. 
The empirical results presented in \cite{makridakis_2018} stress the need for objective and unbiased ways to test the performance of forecasting methods, while \cite{dacrema_2019} investigates recommender systems and finds that less progress is made than what is reported.
\cite{henderson_2018} investigate reinforcement learning and illustrates several practices that lead to poor reproducibility.
\cite{werner2024reproduce} studies how to extend machine learning research by reexecuting, reimplementing, and evaluating an algorithm on another dataset.
\cite{arvan2022} reproduce eight NLP papers based on open code and data at a success rate of 25\% and propose that the complete experiment is shared as self-contained and executable artifacts.
\cite{gundersen2022sources} provides an overview of 41 design decisions that could be sources of irreproducibility. 

The purpose of this study is to shed light on what is important in ensuring the reproducibility of artificial intelligence research. 
Evidence is found by attempting to reproduce 30 AI studies, documenting the problems encountered during reproduction of them, and analyzing which of the problems lead to irreproducible studies. 
Our main contributions are:
\begin{itemize}
    \item A systematic replication study of 30 AI studies that identifies 20 different problem types and their correlation with irreproducibility. 
    \item Empirically establishing the importance of sharing both code and data to ensure the reproducibility of AI research. We also found that the quality of the data documentation has a higher correlation with reproducibility than the quality of the code documentation.
    \item We estimate the reproducibility of AI research to be 33\% when data from the original experiment are available and 86\% when both code and data are available.
\end{itemize}

\begin{table*}
\caption{Overview of the results from the reproducibility studies where Compl. is Completed, Ident. is Identical, Cons. is Consistent and Fail. is Failures. Results of the reproducibility studies can be Success (S), Partial Success (PS), Failed (F), No Result (NR), and Not Started (NS). Paper with ID 23 through ID 30 were R1 studies that were not started (NS).
}
\begin{center}
\begin{tabular}{lccccccccccc}
\toprule
Reference & Type &  Time &  Total & Compl. & Ident. & Cons. & Fail. &       Cause & Result \\
\midrule
\cite{alexe2012measuring}      &   R4 &    40 &     18 &    22\% &     0\% &   22\% &    0\% &        Time &     PS \\

\cite{chen2016generalized}     &   R3 &    40 &      4 &    50\% &    25\% &     0 &   25\% &        Time &     PS \\
\cite{li2012development}       &   R3 &    40 &     46 &   100\% &    37\% &   37\% &   26\% &           0 &     PS \\
\cite{saad2012blind}           &   R4 &    25 &     10 &    10\% &     0\% &   10\% &    0\% &        Code &     PS \\
\cite{li2011cooperatively}     &   R3 &    40 &     21 &    33\% &     0\% &   19\% &   14\% &        Time &     PS \\
\cite{guha2011learning}        &   R3 &    40 &     14 &     7\% &     0\% &    0\% &    7\% &        Time &      F \\
\cite{zeiler2014visualizing}   &   R3 &    40 &     20 &     0\% &     0\% &    0\% &    0\% &        Time &     NR \\
\cite{jia2016isuc}             &   R3 &    22 &      1 &   100\% &     0\% &  100\% &    0\% &           0 &     PS \\
\cite{akay2012modified}        &   R3 &    40 &     68 &    12\% &     0\% &    1\% &   10\% &        Time &     PS \\
\cite{peng2012rasl}            &   R4 &    10 &     31 &    77\% &     0\% &    0\% &   77\% &        Code &      F \\
\cite{liu2015classification}   &   R3 &    40 &      6 &    17\% &     0\% &    0\% &   17\% &  Data, Time &      F \\
\cite{ordonez2016deep}         &   R4 &    20 &      4 &    50\% &     0\% &   25\% &   25\% &        Code &     PS \\
\cite{goferman2012context}     &   R3 &    40 &      2 &   100\% &     0\% &    0\% &  100\% &           0 &      F \\
\cite{le2014distributed}       &   R3 &    40 &      3 &     0\% &     0\% &    0\% &    0\% &        Time &     NR \\
\cite{chen2016xgboost}         &   R4 &    40 &      4 &    50\% &     0\% &   50\% &    0\% &        Time &     PS \\
\cite{zhang2014facial}         &   R3 &    40 &     12 &     0\% &     0\% &    0\% &    0\% &        Time &     NR \\
\cite{chen2014deep}            &   R4 &     8 &     16 &    37\% &    19\% &   19\% &    0\% &        Code &      F \\
\cite{huang2014semi}           &   R3 &    40 &     38 &    13\% &     0\% &    0\% &   13\% &        Time &      F \\
\cite{li2014deepreid}          &   R3 &    22 &      6 &     0\% &     0\% &    0\% &    0\% &        Text &     NR \\
\cite{jia2016deep}             &   R3 &     8 &     10 &     0\% &     0\% &    0\% &    0\% &        Text &     NR \\
\cite{rodriguez2014clustering} &   R4 &    33 &      7 &    86\% &    86\% &    0\% &    0\% &        Text &      S \\
\cite{donahue2014decaf}        &   R3 &    40 &      4 &    25\% &     0\% &    0\% &   25\% &        Time &      F \\

\bottomrule
\end{tabular}
\label{tab:results_overview}
\end{center}
\end{table*}

\small{
\begin{table*}[h]
\caption{
Problem ID, the source of the problem, description of the problem, how many times we encountered the problem, the true positive rate (TPR) and the weights of a logistic regression performing a binary classification on reproducibility result.
}
\begin{center}
\begin{tabular}{cclccc}
\toprule
    ID & Source &  Description of Problem & \# & TPR & $w_i$ \\ 
    \midrule
    P1 & Code & Method code is shared, but not experiment code. & 5 & 0.2 & 0.16 \\ 
    P2 & Code & Method code is  shared but does not cover the entire method.& 1 & 0.0 & 0.12 \\ 
    P3 & Code & Poor documentation of code.& 1 & 0.0 &0.12 \\ 
    P4 & Code & Experiment setup not not described or differ from article.& 2& 0.0 & 0.47 \\ 
    P5 & Code & Not clear which version of the code is used for the experiments.& 3 &  0.0 & 0.44 \\ 
    P6 & Code & Code is shared in a compiled or non-inspectable form. & 2& \textbf{1.0} & -0.4 \\ 
    P7 & Code & Random seeds and random number generators not specified. & 4 & 0.0 & \textbf{0.88} \\ 
    P8 & Article & Ambiguous description of method.& 5 & 0.6 & -0.28 \\ 
    P9 & Article & Ambiguous description of the experiment.&7 & 0.7 & -0.27  \\ 
    P10 & Article & Ambiguous description of the implementation of the method or experiment.& 12 &0.6 & -0.15 \\ 
    P11 & Article & Multiple methods for solving a sub-task are mentioned but not specified. & 2 & 0.5 & 0.11 \\ 
    P12 & Article & Hyperparameters are shared online, but differ from those used in experiment.& 1 & 0.0 & 0.12\\ 
    P13 & Article & Experiment- or hyper-parameters are not shared.& 6 & 0.3 & 0.39 \\ 
    P14 & Article & The article contains an error.& 1 & 0.0 & 0.26 \\   
    P15 & Data & Mismatch between dataset specified and version of it found online. & 5 & \textbf{1.0} & \textbf{-1.00} \\ 
    P16 & Data & A subset of a dataset is used, but which exactly is not specified. & 1 & \textbf{1.0} & -0.20\\   
    P17 & Data & The preprocessed or augmented version of a dataset is not shared.& 4 & 0.8 & -0.06\\   
    P18 & Data & How dataset is partitioned into training, validation, and test is not described.& 4 & \textbf{1.0} & \textbf{-0.82} \\
    P19 & Results & Results are presented in a way that makes a comparison hard. & 2 & 0.5 & 0.29 \\
    P20 & Resources & Lack of access to hardware or software needed to conduct the experiment.& 1 & \textbf{1.0} & -0.36\\
\bottomrule
\end{tabular}
\end{center}
\label{tab:problem_types}
\end{table*}
}

\section{Selecting and Retrieving Articles}
In total, 30 articles were included in this study. 
We included ten articles from each of the years 2012, 2014, and 2016. 
Using scopus.com, 
a search was performed for each year for empirical articles in AI, and the results were ranked according to the number of citations.
Then, the ten most cited articles from each year were selected for inclusion.
An initial check revealed that not all the articles returned by Scopus were empirical studies. 
These articles were excluded from the study and replaced by the empirical studies that followed next in the ranking.
Therefore, the final list of articles contains the ten most cited articles (as of January 2018) reporting empirical AI research from 2012, 2014, and 2016.
\cite{raff2019step} saw an increase from 63.5\% reproducibility to 84\% after contacting the authors of the original article.
We did not contact the authors as the documentation should contain all necessary information. 

\section{Determining Reproducibility Type} 
Before starting our attempt at reproducing the 30 retrieved articles, we determined their reproducibility type.
\cite{gundersen2022machine} describes four types of reproducibility based on the documentation available to third parties: \emph{R1 Description} where only the research report (scientific article) is available, \emph{R2 Code} where the research report and the code implementing the experiment are available, \emph{R3 Data} where the research report and data used in the experiments are available, and finally \emph{R4 Experiment} where the research report, code, and data are available.
We rely on this classification here.
We only included articles that were of type R4 Experiment and R3 Data. 
All articles for which we found both code and data were classified as R4, while those for which we only found data were classified as R3. 
We did not attempt to replicate articles that relied on closed data, as it was practically impossible due to the workload associated with data collection, restricted access to data sources, and privacy issues.
So, articles that were of type R1 Description and 
R2 Code 
were not reproduced.
Articles of both types are classified as R1 Description. 

\subsection{Code Retrieval}
For a study to be classified as R4 Experiment, the code implementing the method was required to be publicly available for \emph{at least one} of the experiments described in the article.
As such, replicating R4 articles may involve the writing of new code, primarily implementation of experiments. 
We search for the code in three different ways when determining whether a study is of type R1. 
First, we checked whether the research article contained a link to the code. 
Then, if it did not contain a link to the code, we searched the Web using google.com 
twice if necessary: one search containing only the name of the study and another with the name of the study followed by the term "github" to specifically check whether the code was hosted at github.com. 
Finally, we checked the web pages of the main authors of the study.
If an implementation was found, we tried to determine whether it was the original implementation or one done by a third party. 
This was done by checking whether the implementation explicitly mentioned being part of the original study or whether the code maintainer was one of the authors of the original study. 
If either of these requirements were met, the implementation was assumed to be original.
Implementations that were shared in an uninspectable manner, that is, as compiled programs, were included in this study as R3 and not R4 studies.

\subsection{Data Retrieval}
A similar procedure was used when searching for data sets.
For articles relying on data sets, we checked whether the article provided links to web pages hosting the data sets or references to articles that introduced the data set. 
If a link was not provided, we searched for the data set name using the Google search engine. 
If no matches were found, we searched the referenced articles for information on where to find the data sets. 
When determining whether a data set found online was the same as the one used in the study, it was assumed that the data set was the same if the name was identical or if the original article and the page that hosted the data set referenced the same research article.

In some cases, articles will rely on popular data sets in one or more of its experiments but will perform significant preprocessing on the data before using it in the proposed method.
Preprocessing may change the data samples or the composition of the data set resulting in a new data set slightly different from the original. 
However, since studies often use data sets owned by other researchers or institutions, authors may have limited ability to share the new data sets that are preprocessed. 
When encountering studies with data preprocessing where the processed data is not shared, we tried to re-run, and if necessary reimplement, the data preprocessing pipeline. 
The only exceptions to this were studies where the preprocessing had to be reimplemented and involved multiple complex stages, or where the preprocessing required manual editing. 
In these cases, the likelihood of us creating a data set that is identical or equivalent to the original data set was deemed low. 
Studies with these types of preprocessing were considered R3 reproducible, on the grounds that an important part of the methodology used to produce the results needed reimplementation. 

\section{The Replication Study}
A reproducibility study that reproduced the results reported in an article\footnote{https://github.com/AIReproducibility2018} was carried out at the highest reproducibility level possible.
So, if an article was considered to be of reproducibility type R4, a reproducibility study relying on both code and data was attempted. 
Hence, for R4 articles, we did not reimplement the code, except if parts of the code were missing. 
The maximum time we spent on a reproducibility study was 40 hours of focused work.  
Breaks did not count towards the limit. 
The limit was set for practical reasons, and we considered 40 hours a reasonable effort. 
To some extent, we based this decision on the prediction that well-documented studies should be reproducible within this time frame.

Many published articles include more than one experiment, so one reproducibility study could contain several reproducibility experiments. 
We focused on one experiment at a time. 
When deciding which experiment to start with, we emphasized the importance of the experiment to the article, i.e. how much it is discussed, as well as the order in which the experiments were presented. 
In most cases, the first eligible experiment was conducted first. 
Some experiments described in the articles were excluded on the basis of which material was available for exactly that experiment.
For example, when deciding that an article could be reproduced at the R4 reproducibility level, only the experiments in the article covered by the provided method code were considered eligible. 
If after having achieved results for the first experiment and there was still time left of the 40 hours, we move on to the next eligible experiment.

Counting the number of experiments in an article and distinguishing between the experiments was difficult at times. 
Since different articles used the term experiment in different ways, there is a need for a clear definition. 
The definition we rely on is that one experiment is one method run on one data set or function.
When running multiple methods or using multiple data sets, these are considered to be multiple experiments, even if the original article does not specify them as such.

For R4 articles, we implemented methods and experiments from scratch. 
In these cases, we chose to use the same programming language and third-party libraries that were used in the original study. 
If the programming language was not specified, we chose a suitable programming language. 
When the use of a third-party library was mentioned, we attempted to use the same library and version as specified in the article. 
However, if the library mentioned was unavailable, a substitute library was used. 
We also used third-party libraries not mentioned in the original article in our implementation in cases where this was considered practical. 
For example, in cases where a study used a known algorithm as part of its method but did not describe how this algorithm was implemented, we relied on a third-party implementation. 
Whenever a random number generator was used, we explicitly set the seed in the code.
All results produced by a method, not just metrics and result aggregates, were written to file in our study.

Some reproducibility studies were aborted before all experiments were attempted, or the time limit was reached. 
We also aborted reproducibility studies where the hardware demand exceeded what was available to us or in cases where a reproduction attempt reached a situation where it was impossible to get any results within the remaining time of the reproducibility study.
Such situations were recorded as spending all the time (40 hours) even though this was not the case in practice.
We used personal computers and a high-end GPU cluster to execute experiments.

Figure \ref{fig:reproducibility_experiment} shows how a reproducibility study relates to the original study reported in an article. 
The research design, which is part of the article, describes how the experiment should be conducted, and the experiment is executing tailor-made code on a dataset utilizing some ancillary software, i.e. libraries, frameworks, and operating systems, on some computer hardware. 
The outcome of the computational experiment is analyzed and a conclusion is reached.
Many assumptions must be made when conducting an R3 reproducibility study, as the article cannot possibly cover all design decisions. 
The figure also emphasizes that the code, hardware, and ancillary software differ from the original experiment, as does the outcome for all of the experiments reported in the article. 
As one of the conclusions differs, the study is deemed a Partial Success.

\section{Classifying Results}
Analyzing whether replications are successful requires that the results be grouped together.
There are two levels of results to consider: 1) the result produced by reproducibility \emph{experiments}, of which there can be many per study, and 2) the results of reproducibility \emph{study}, which is the aggregate of these reproducibility experiments. 
The distinction between the results reported in the study as a whole and each individual experiment is usually not discussed. 
For example, \cite{gundersen2022sources} does not explicitly distinguish between the study and the experiments.
However, their definition of a reproducibility experiment is compatible with an individual experiment but not with the whole study, which potentially consists of many experiments.
\cite{raff2019step} distinguish between these and considers a study to be reproduced if more than 75\% of the experiments support the conclusion. 

\subsection{Experiment Result Classification}
The classification of each of the individual reproducibility experiments reported in an article is addressed first.
We follow the terminology used by \cite{gundersen2022sources} in which they distinguish between \emph{outcome reproducible} and \emph{analysis reproducible}.
In cases where the reproduced outcome of an individual reproducibility experiment is \emph{identical} to the original experiment, the reproducibility experiment is considered outcome reproducible.
However, in many cases, the reproduced outcome will not exactly match the original values, but the same analysis of the different outcomes will lead to the same conclusions. 
Reproducibility experiments that generate different outcomes than the original study are analysis reproducible if the analysis is \emph{consistent} with the original analysis and results in the same conclusion. 
If the result of a reproducibility experiment is classified as neither identical nor consistent, it is classified as \emph{failed}, and the reproducibility experiment is negative; the result is not reproduced. 
Hence, there are three possible outcomes for each reproducibility experiment: 1) identical, 2) consistent, and 3) failed.

\subsection{Study Result Classification}
The result of a reproducibility study is decided on the basis of the aggregate result of all individual reproducibility experiments. 
A reproducibility study is defined as \emph{Success} if all of the experiments are either outcome or analysis reproducible, which means that each of the reproducibility experiments is identical or consistent with the original experiment. 
If there is a mixture of identical, consistent and failed reproducibility experiments, the reproducibility study is classified as \emph{Partial Success}.
If all reproducibility experiments of a study failed, the reproducibility study is classified as \emph{Failure}.
If none of the reproducibility experiments was completed successfully, the result of the reproducibility study is \emph{No Result}. 
The R1 reproducibility studies that we were unable to conduct because we did not have access to relevant data were classified as \emph{Not Started}.

What constitutes a Partial Success?
A study for which only one of the experiments were identical or consistent is classified as Partial Success even though the study reported many experiments. 
There are many sources of irreproducibility \cite{gundersen2022sources}, some of which are hard to control, such as variability introduced by different hardware or software stacks.
This could mean that the researchers conducting the original study could reproduce the results they reported consistently while we, a third party, were not able to.
So, a Partial Success provides evidence of the experiments being documented so well that third parties are able to repeat them.

\section{Results}
\begin{figure*}[t]
     \centering
     \begin{subfigure}[b]{0.49\textwidth}
         \centering
         \includegraphics[width=\textwidth]{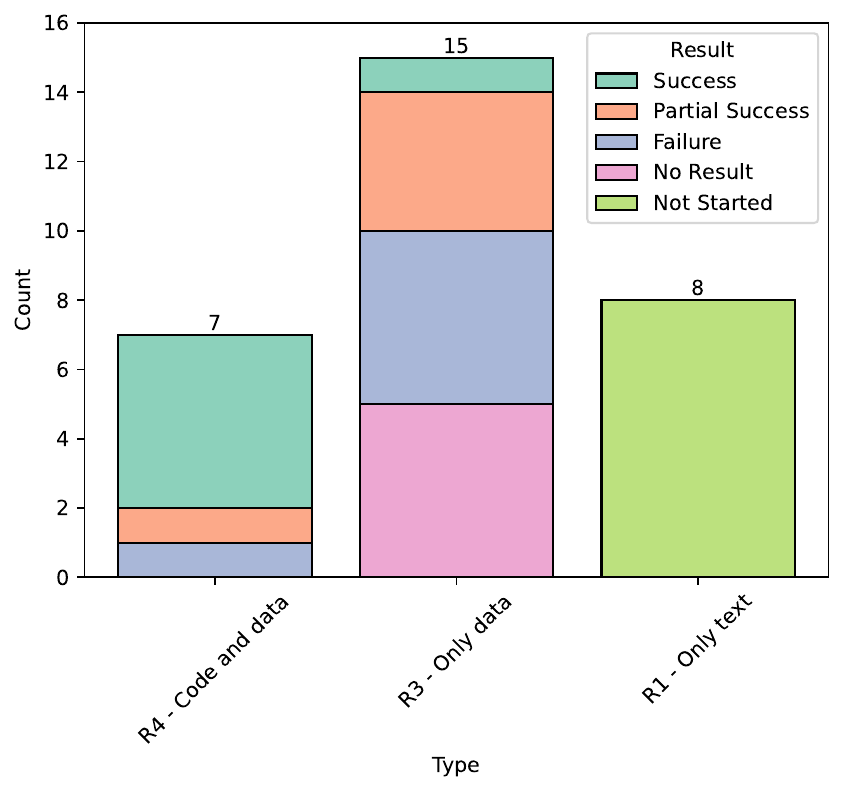}
         \caption{Results per reproducibility type.}
         \label{fig:type-hist}
     \end{subfigure}
     \hfill
     \begin{subfigure}[b]{0.49\textwidth}
         \centering
         \includegraphics[width=\textwidth]{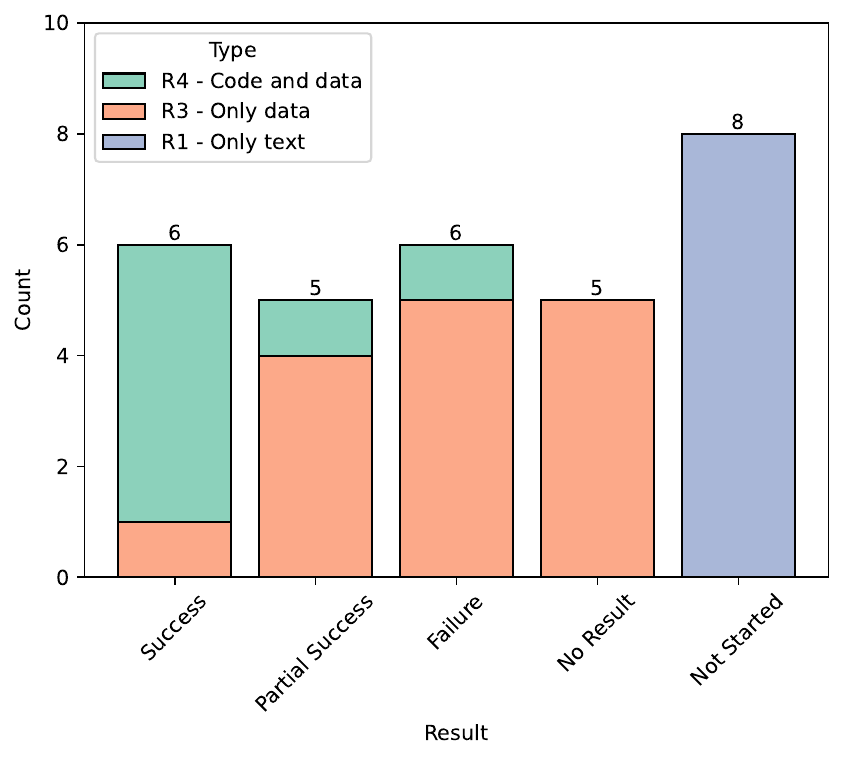}
         \caption{Reproducibility types per result.}
         \label{fig:results-hist}
     \end{subfigure}
        \caption{Overview of the results. }
        \label{fig:histograms}
\end{figure*}

Table \ref{tab:results_overview} provides an overview of the 22 studies we attempted to replicate.
The table shows the type of reproducibility, the amount of time spent on reproducibility studies, the total number of experiments in a study, the percentage of experiments that were completed and the percentage of results that were identical to, consistent with, or different from the original study. 
The reason for not completing all experiments in a study is presented next, and finally, the results of the reproducibility studies are provided.
Eight of the 30 articles were rejected because they required us to collect new data (R1 Description type) or use special hardware, and these were \cite{cheng2016learning}, \cite{meng2016mllib}, \cite{shin2016deep}, \cite{silver2016mastering}, \cite{graves2014towards}, \cite{yin2013robust}, \cite{zhang2012multi}, \cite{zhang2012single}.
Seven of the 22 articles were of reproducibility type R4 where both data and code were made public, while the remaining 15 articles were of reproducibility type R3 where only data were made available. 

We completed all experiments for three of the studies, and we spent all available time (40 hours) for 12 studies without completing all experiments, which means that 40 effective working hours is not enough time to replicate a study -- surprisingly also for R4. 
Missing code was the second most common cause for not completing a study.
This meant that even though parts of the code were made available, not all code was, so we had to reimplement parts of the codebase, which, of course, is time-consuming. 
This was the reason for not completing four of the studies. 
For three of the studies, poor descriptions were the reason for not completing them, as it was not clear what to do. 

Figure \ref{fig:histograms} provides an overview of the results of the reproducibility studies. 
As Figure \ref{fig:type-hist} shows, of the seven R4 studies, five were reproduced successfully and one partially, while only one of the 15 R3 studies was reproduced successfully and four partially, which means that ten were failures (5) or not completed (5). 
Figure \ref{fig:results-hist} illustrates well that the sharing of code and data leads to successful replications. 
Studies that only share data are the most prevalent for partial successful replications, but more fail and end with no result. 
We reproduced 11 out of 22 studies successfully (six) or partially (five), while we did not reproduce six and obtained no result for five, leading to a success rate of 50\% for those we started. 

In Table \ref{tab:problem_types}, we provide an overview of the 20 different types of problem that we encountered when trying to reproduce the 22 studies.
The problem types represent five different sources of problems: 1) source code, 2) article, 3) data, 4) results, and 5) resources. 
The code and the article were the most prevalent sources of problems, as they were the source of seven problem types each, while the data was the source of five of the problem types.
The reporting of results was the source of one problem type, and the lack of access to resources, such as the computing infrastructure, was the source of one problem type. 
The 20 problem types were observed in total 69 times (sum of column \# in the table).
Ambiguous descriptions of the method (P8), experiment (P9), and implementations (P10) were encountered totally 24 times. 
More than half of the studies (12) described the implementations ambiguously (P10).
Precisely describing how an experiment is conducted will in most cases require careful documentation and often longer descriptions than what is possible in academic literature, which emphasize the importance of sharing code; code helps to disambiguate the description.

\begin{table}[t]
\caption{Results grouped by reproducibility type: Success (S), Partial Success (PS), Failure (F), No Result (NR), Inclusive Success (IS) and Inclusive Failure (IF).}
\begin{center}
\begin{tabular}{ccccccc}
    \hline
    Type & S & PS & F & NR & IS & IF\\ 
    \hline
    R3 & 1 & 4 & 5 & 5 & 5 & 10 \\
    R4 & 5 & 1 & 1 & 0 & 6 & 1 \\
\hline
\end{tabular}
\label{tab:results_per_reproducibility_type}
\end{center}
\end{table}

The sample size of this replication study is small. 
It includes 30 studies in total, however, eight of them were rejected, which means our analysis includes only 22 studies. 
To increase the robustness of the findings, we group the results into \emph{Inclusive Success}, which is the sum of Successes (S) and Partial Successes (PS), and \emph{Inclusive Failure}, which is the sum of Failure (F) and No Result (NR).
Table \ref{tab:results_per_reproducibility_type} provides an overview of the results for the two types of reproducibility studies. 
The main findings is that sharing both code and data increases the chance of reproducing results immensely. 
For R3 studies where only data is shared, the probability of a successful reproducibility (IS) study is only 33\% (five out of ten studies). while for R4 studies where both data and code is shared, reproducibility increases to 86\% (six out of seven studies). 

To determine which types of problem are most relevant for distinguishing between studies that are reproducible and those that are not, we performed a binary classification using logistic regression, which is a simple and efficient way of doing this. 
We classified the studies into whether they were reproducible or not with problem types as features.
With 11 samples of each class, the dataset is balanced.
Due to the small sample size and the aim of determining the significance of the features, we dropped the validation and test sets. 
The accuracy of the classification was $0.91$ and a bias of $0.20$, so the logistic regression was clearly capable of distinguishing between reproducible and non-reproducible studies. 
As can be seen in Table \ref{tab:problem_types}, the three most significant features ($\|w_i\| > 0.8$) are P15 which is a mismatch between the dataset specified and the one found online, P7 which is that random seeds or random number generator are not specified, and P18 is when the dataset description lacks information how it is divided into training, validation and test sets. 
The source of both P15 and P18 is the data, while the source of P7 is the code.

It is interesting to establish which problems characterize irreproducible studies. 
To analyze this, we calculated the true positive rate (TPR) for each type of problem.
The result is listed in the TPR column of Table \ref{tab:problem_types}.
A type of problem with a TPR of $1.0$ is only encountered in irreproducible studies, while a TPR of $0.0$ only characterizes reproducible studies. 
As our goal is to find out what causes irreproducibility, we focus on the five types of problem with a TPR of $1.0$.
The source of three of these problem types, P15, P16, and P18 is the data, while code being shared in compiled form (P6) and lack of access to specialized hardware or software are the two others (P20).
Notably, P17, which is the fourth of the problem types with data as the source of problems, has a TPR of 0.8, indicating that problems with documenting the data work are the most important source of irreproducibility. 
This is further emphasized by the fact that both the weights of logistic regression and TPR indicate that problem types P15 and P18 characterize irreproducible research. 
The source of these is poor data documentation.

The most surprising result might be the fact that the sharing of code is so effective in ensuring reproducible studies, as 86\% of R4 studies are reproduced, while problem types related to code are not indicative of irreproducible research. 
Problems related to the quality of the documentation of code, whether it is partially missing or not versioned, are not important for successful replication as long as code is shared. 
Only when code is shared in an uninspectable manner is it a source of irreproducibility.
The fact that both studies that shared code in an uninspectable form were not reproduced might indicate that sharing code in this way is a mark of sloppy research; the research poses as open, as the code is shared, but actually it is closed, as the code is uninspectable.
Sharing uninspectable code is like not sharing code at all.

\section{Discussion}

Clearly, sharing code and data will enable others to carry out the experiment, as is emphasized by \cite{arvan2022,stodden2018empirical,collberg2016repeatability} who, respectively, were able to \emph{repeat} 25\%, 26\%, and 32\% of the selected studies by executing the open experiments.
However, unless those reproducing the results check and understand the code, the results might not reflect a true finding, even in a restricted sense where findings only apply for the selected datasets.
The code may not implement the methodology or experiment described in the article \cite{raff2023siren}, or the code or the ancillary software may contain errors \cite{thomson2024common}.
Sharing self-contained and executable artifacts will simplify the verification that running the code produces the reported result, but is not sufficient for reproducibility; repeatability is not reproducibility. 
Our concern is reproducibility.

Sharing both code and data is important, as illustrated by the reproducibility checklists of the top AI conferences. 
We reviewed the reproducibility checklists for IJCAI 2024, AAAI 2024, and NeurIPS 2023, and therefore also the ICML reproducibility checklist, which was based on the NeurIPS checklist. 
All checklists require researchers to share code and data, but it is not required.
However, only 46\% of the articles claim to open-source their code \cite{magnusson-etal-2023-reproducibility}.
One could ask whether open code and data should be mandated and that exceptions should be argued for before a paper is accepted for publication. 
Then, poor reasons for not sharing code and data could lead to summary rejection. 
The reproducibility checklists also request that the hyperparameters of the experiments be shared. 
Surprisingly, according to this study, sharing experiment parameters and hyperparameters is not associated with successful replications.
AAAI, NeurIPS and ICML emphasize the need for sharing not only the method code but the experiment code as well.
This reduces ambiguity, but our study finds that this does not correlate with reproducibility.
Sharing \emph{some} code is sufficient. 
IJCAI's reproducibility checklist contains only a few items, but gives extra focus to describing the computing infrastructure. 
The computing infrastructure is not found to be important for reproducibility. 
While NeurIPS and ICML specify that important details of experiments should be described in the article even if code is shared, the AAAI checklist details that method, experiment and data should be described in detail, which is best practice according to this study. 

Not surprisingly, ambiguous descriptions of the method, experiment, and implementation will affect the reproducibility of a study (P8, P9, and P10). 
However, whether a study is reproducible or not depends on the information left out.
Some information is, of course, more important to relay than others. 
This indicates that researchers might not be able to distinguish between which information is important and which is not.
As we did not document this in more detail, our study does not provide further insight, but clearly this could be a subject of future research.

It is natural to compare our approach to the approach used by \cite{raff2019step}, which was a study carried out by a person over several years where 
The articles were selected according to their relevance to the JSAT library \cite{raff2017jsat}. 
The sample is biased towards statistical methods. 
Also, the knowledge required for implementing these methods is narrow in one sense, and the knowledge of previous methods could increase the success of implementing the next method, which, of course, is a positive side-effect of experience.
However, this could mean that the estimated reproducibility of 63.5\% for machine learning probably is an overestimation. 
Our sample of studies has a different bias, as we retrieved studies covering AI that were highly cited.
The sample is broader in the topics covered. 
However, retrieving the most cited articles could contribute to an overestimation, as many have found these studies valuable and cited them. 
Furthermore, our study was conducted by a small team with less knowledge in the very broad scope of the retrieved studies, probably leading to less positive effects of experience in reproducing previous studies. 
\cite{raff2019step} does not rely on any source code released by the authors of the original articles; it is not clear on which data the reproducibility experiments rely. 
Therefore, the studies could be of type R1 description or R3 data.
In addition, \cite{raff2019step} mentions that a study is deemed reproducible if 75\% of the claims in the article are supported, which is less restrictive than \emph{Success} and more restrictive than \emph{Partial Success}.
However, it is not clear from the paper how the claim is interpreted or how a claim is evaluated.

\section{Limitations}
\label{sec:limitations}
This research has several limitations. 
First, we estimate the reproducibility of AI research to be 50\% for articles that rely on open data. 
However, this number is highly uncertain. 
We reproduce well-cited research, which could lead to overestimation. 
These articles were the most cited empirical studies in AI when the selection was done in 2018. 
Currently, the most cited article is cited over 34 000 times, and the least cited article is cited just below 300 times (mean citation is 4499 and median citation is 1666). 
According to \cite{van2014top}, having 100 or more citations makes a study among the top 2.6\% most cited articles. 
Hence, the articles part of this study can arguably be considered highly cited articles, which means that the selection of articles is biased towards high-quality research, indicating that our estimation of the reproducibility is an overestimation. 
However, our estimate could be an underestimation. 
\cite{pineau2021improving} found that the introduction of the reproducibility checklist in ICML and NeutIPS made publicly shared code more prevalent.
However, this study contains articles from before reproducibility checklists were introduced at all the top AI and machine learning conferences.
Although their finding has not been verified to apply generally, one would assume that reproducibility has increased because of the introduction of reproducibility checklist, especially as these reproducibility checklists focus on making code and data publicly available.
Our estimate of reproducibility is only for papers that share data or code and data, which means that it estimates just a subset of all reproducibility types. 
This probably also points towards an overestimate. 

Second, the introduction of reproducibility checklists has probably on average increased the prevalence of code and data sharing given how this was the case when introduced for ICML and NeurIPS \cite{pineau2021improving}.
This means that the number of R1 vs. R3 vs. R4 studies reported here probably does not reflect the actual current state - at least for the top conferences that have introduced such checklists. 

Third, we only allotted 40 hours per article, which clearly was not enough to reproduce all the experiments of the studies reported here. 
All experiments were completed for only three of the 22 studies that we attempted to reproduce, and
for only one of these, less than 40 hours were used to complete the experiments. 
Surprisingly, only articles that did not share code were 100\% completed. 

Fourth, we did not get a result (NR) for articles \cite{li2014deepreid} and \cite{jia2016deep}, as the articles did not provide enough information for us to reproduce them (Text as Cause in Table \ref{tab:results_overview}).
The root cause could be that the article lacked detail or that we lacked the knowledge required to properly reproduce the articles. 
Could one expect that anyone (in AI) should be able to reproduce the results of an article listed as AI and in 40 hours?
We only reproduced all experiments reported in three of the 22 articles in 40 hours or less, so clearly this could not be expected. 
The allotted time was not enough even for R4 studies that shared both code and data. 
Researchers might spend many years understanding a concept and a long time designing and implementing experiments. 
It takes time to both deeply understand a study and reimplement it even when code and data are shared.

Fifth, the sample size is small, since the study only attempts to reproduce 22 articles, which means that all estimates are highly uncertain. 
Increasing the sample size by including more articles would reduce uncertainty, but due to resource constraints, this was not possible. 
Finally, although we have tried our best to avoid it, our implementations can, of course, contain errors and bugs, and we could have misunderstood concepts we thought we understood. 
Hence, our experiments and findings might, of course, have similar types of issue as all other research and even the problem types that were identified and discussed in this article. 

\section{Conclusion}
This study emphasizes the effectiveness of open science. 
The sharing of both code and data publicly is extremely important to ensure the reproducibility of AI studies, since 86\% of the studies that shared both code and data were reproduced while only 33\% of the studies that only shared data were reproduced. 
This study also establishes that documenting the data work is extremely important to ensure reproducibility, while documenting code is less so, as long as code is shared. 
Not sharing data is an obstacle for third parties to reproduce studies, as it requires third parties to collect new data. 
In many cases, this cannot be avoided, that is, when protecting privacy is of importance, which is typically the case in medical studies. 

Although requiring all studies published in a venue to share both code and data could be too strict, the default expectation should be that both code and data are shared. 
Researchers should be required to argue why they cannot share data or code. 
If the reasons for not sharing are poor, then an idea might be to summarily reject the research.
This would be a clear incentive for researchers to conduct open science and enable others to build on their findings. 
Our study clearly suggests that this will improve the reproducibility of AI research.
It will also simplify generalizing to new data \cite{werner2024reproduce}.

These results have important implications for current research trends.
A worrisome trend in research on large language models (LLMs) is that experiments are conducted on LLMs for which neither code nor training data are available, such as those developed by large technology companies.
It means that the results could be (almost) impossible to independently reproduce, whether it is emergent behaviors \cite{wei2022emergent} or something else. 
Importantly, third parties will not be able to investigate exactly what is the cause of different effects when the code and data are not available.

%

\bibliography{references}

\end{document}